# Hybrid Henry Gas Solubility Optimization Algorithm with Dynamic Cluster-to-Algorithm Mapping for Search-based Software Engineering Problems

Kamal Z. Zamli[1*], Md. Abdul Kader[1], Saiful Azad[1] and Bestoun S. Ahmed[2]

[1]*Faculty of Computing, College of Computing and Applied Sciences*
*Universiti Malaysia Pahang, 26300 Kuantan, Pahang, Malaysia*
*Email: kamalz@ump.edu.my[*], kdr2k10@gmail.com, saifulazad@ump.edu.my*

[2]*Department of Mathematics and Computer Science, Karlstad University, 65188 Karlstad, Sweden*
*Email: bestoun@kau.se*

**Abstract**
Henry Gas Solubility Optimization Algorithm (HGSO) is a recently developed population-based meta-heuristic algorithm in the literature. One notable feature of HGSO is that the algorithm divides its (single) population into a set of clusters that are individually mapped to an independent HGSO with its parameter settings (as well as its local best). At a glance, having multiple independent HGSO serving the given clusters in the population can definitely boost exploration (i.e., in terms of roaming the new potential region in the search space for better solution alternatives). However, a closer look reveals two main limitations. Firstly, HGSO-to-cluster mapping is statically defined. To be specific, the defined HGSO-to-cluster mapping does not consider its adaptive performance for the subsequent iteration. Secondly, HGSO implementation ignores the opportunity for hybridization with other meta-heuristic algorithms. With hybridization, one can compensate the limitation of a host algorithm with other algorithms' strength. Best results in the literature have often been associated with hybridization. Addressing these limitations, this paper proposes the development of Hybrid HGSO (HHGSO). Taking HGSO as the host algorithm, HHGSO is hybridized with four recently developed meta-heuristic algorithms, including Jaya Algorithm (JA), Sooty Tern Optimization Algorithm (STOA), Butterfly Optimization Algorithm (BOA) and Owl Search Algorithm (OSA). The individual mapping of each algorithm is made dynamic based on penalized and reward adaptive probability. Comparative performance of HHGSO with the aforementioned algorithms is conducted with a well-known Search-based Software Engineering (SBSE) problem involving team formation problem. Additionally, the defined hybridization approach has also been adopted as a hybridization template for solving the combinatorial test generation problem with the same meta-heuristic algorithm combinations. Comparative performance is also undertaken against recently developed hyper-heuristic algorithms involving Exponential Monte Carlo with Counter, Modified Choice Function, Improvement Selection Rules, and Fuzzy Inference Selection. Our results indicate that the HHGSO hybridization has usefully improved the performance of the original HGSO and gives superior performance against the given competing algorithms.

## 1. Introduction

The meta-heuristic algorithm can be seen as a template for solving general optimization problems. Guided by a given (minimization or maximization) objective function, every meta-heuristic algorithm provides a specific mechanism to explore (i.e., roaming the new potential region in the search space for better solution alternatives) and to exploit (i.e., manipulating the search space in the vicinity of the known best) the given search space efficiently. To ensure good performance, the exploration and exploitation need to be properly balanced during the actual search process (i.e., considering the search space contour). Given the diverse source of inspirations (e.g., physical-based [1], swam-based [2], natural evolution-based [3]) for each meta-heuristic algorithm, the exploration and exploitation process in each meta-heuristic algorithm can be significantly different.

---
[*] Corresponding Author

To date, the best classification of existing meta-heuristic algorithms with clear group demarcation relates to a *single solution* and *population-based* algorithms. As the name suggests, single solution based meta-heuristic algorithms use a single solution model. In each iteration, the same solution is updated until the iteration ends. The advantage of a single solution-based model is that it is memoryless. As a result, the execution overhead of single solution-based algorithms is often low. However, single solution-based algorithms often have slow convergence as the algorithms rely on a single point exploration. On the same note, the performance of single solution-based meta-heuristic algorithms is also sensitive to its initial search point. Some examples of single solution-based meta-heuristic algorithms include Simulated Annealing (SA) [4], Guided Local Search (GLS) [5], Variable Neighborhood Algorithm (VNS) [6], Threshold Accepting Method (TA) [7] and Tabu Search (TS) [8].

On the other hand, population-based meta-heuristic algorithms exploit a set of solution candidates (i.e., memory) via its population. Throughout the search process, every candidate solution in the population is updated until the iteration ends. The main advantage of the population-based meta-heuristic algorithm over the single population-based counterpart relates to exploration. Given that each population may be residing in a different region of the search space, convergence of population-based is much faster than single population-based meta-heuristic algorithms. However, the execution overhead of population-based meta-heuristic algorithms is often higher. Some examples of population-based meta-heuristic algorithms include Jaya Algorithm (JA) [9], Sooty Tern Optimization Algorithm (STOA) [10], Butterfly Optimization Algorithm (BOA) [11], Owl Search Algorithm (OSA) [12], Henry Gas Solubility Optimization Algorithm (HGSO) [13], Manta Ray Foraging Optimization (MRFO) [14], Water Strider Algorithm (WSA) [15], Artificial Electric Field Algorithm (AEFA) [16], Equilibrium Optimizer (EO) [17], Side-Blotched Lizard Algorithm (SLA) [18], Political Optimizer (PO) [19], Poor and Rich Optimization Algorithm (PROA) [20], and Tunicate Swarm Algorithm (TSA) [21].

Henry Gas Solubility Optimization Algorithm (HGSO) is a recently developed population-based meta-heuristic algorithm in the literature. The notable feature of HGSO that stands out against other competing meta-heuristics is that the whole population (e.g., pop [0] till pop [k] in Figure 1) is divided into a set of N clusters. Each cluster is then mapped to an independent HGSO with its parameter controls as well as its own local best. At a glance, having more than one independent HGSO in many sub-clusters can boost exploration. However, a closer look reveals two main limitations. Firstly, the HGSO-to-cluster is statically mapped (i.e., at any instance of iteration, the same mapping is used). Here, the static mapping does not consider the adaptive performance of each HGSO-to-cluster mapping for the subsequent iteration. Secondly, HGSO implementation also ignores the opportunity for hybridization. With hybridization, one can compensate for the limitation of a host algorithm with the strength of others. Best results in the literature have often been associated with hybridization [22-24]. Rather than having one type of meta-heuristic-to-cluster-mapping, many types of meta-heuristic-to-cluster-mapping can be introduced by using more than one algorithm. Additionally, the meta-heuristic-to-cluster-mapping can be made adaptive and dynamic based on the need for the current search process.

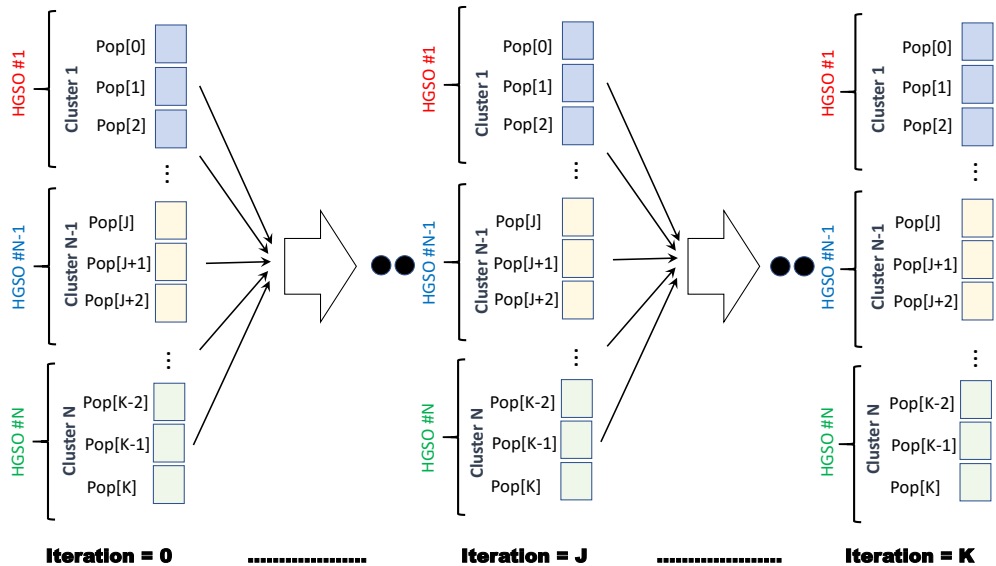

Figure 1. HGSO-to-Cluster Mapping

Given the aforementioned prospects, this paper proposes the development of Hybrid HGSO (HHGSO). The main contributions of this work are as follows:
- A new HGSO variant, called HHGSO, exploiting four recently developed meta-heuristic algorithms (apart from itself as host) including Jaya Algorithm (JA) [9], Sooty Tern Optimization Algorithm (STOA) [10], Butterfly Optimization Algorithm (BOA) [11] and Owl Search Algorithm (OSA) [12]. Here, the individual mapping of the algorithm to the cluster is made dynamic based on penalized and reward adaptive probability.
- Performance comparison of HHGSO with other state-of-the-art algorithms (e.g., the original HGSO, Jaya, Sooty Tern Optimization Algorithm (STOA), Butterfly Optimization Algorithm (BOA) and Owl Search Algorithm (OSA) as well as recently developed hyper-heuristic algorithms (e.g., Exponential Monte Carlo with Counter, Modified Choice Function, Improvement Selection Rules, and Fuzzy Inference Selection ) for two Search-based Software Engineering problems involving team formation problem and combinatorial test suite generation.

The paper is organized as follows. Section 2 describes the related work on the hybridization of meta-heuristic algorithms. Section 3 presents the original HGSO. Section 4 elaborates on the proposed HHGSO. Section 5 presents our empirical evaluation along with our research questions. Section 6 discusses our experimental observations related to all the research questions. Finally, section 7 gives our concluding remark along with the scope for future work.

## 2. Related Work on Meta-Heuristic Hybridization

Without any knowledge about the nature of the optimization problem at hand, combining the strengths of different meta-heuristic algorithms (as well as compensating the limitation of host algorithm), belonging to different classes of implementations, may increase the probability of success of the overall search process. Specifically, meta-heuristic hybridization involves combining or grouping two or more meta-heuristic algorithms to form a new hybrid algorithm. It should be noted that we differentiate hybrid algorithms from ensemble algorithms. Ensemble algorithms involves combining one or more meta-heuristic algorithms with other general artificial based algorithm such as artificial neural network, fuzzy logic, k-mean clustering algorithm and support vector machine to name a few.

To put meta-heuristic hybridization into perspective, consider the different possible grouping for utilizing and combining meta-heuristic algorithms. Owing to its popularity and in line with the scope of the current work, our groping focuses only on the hybridization of population based meta-heuristic algorithms (see Figure 2). The first two (non-hybrid) groups, shown for completeness, are single algorithm single population (SASP) and single algorithm many populations (SAMP) class. These two classes can be ignored from our discussion as we want to focus on hybridization methodologies. The two remaining groups are many algorithms single population (MASP) and many algorithms many populations (MAMP). These two grouping can further be decomposed into low-level hybrids and high-level hybrids based on Talbi [25] and Zamli [24]. Low-level hybrids require changes in the internal structure of the source code to alter certain function(s) of a host algorithm with other meta-heuristic algorithms. Meanwhile, high-level hybrids combine two or more meta-heuristic algorithms as black-box components (i.e., the internals are non-intersecting with the implementation of the host algorithm).

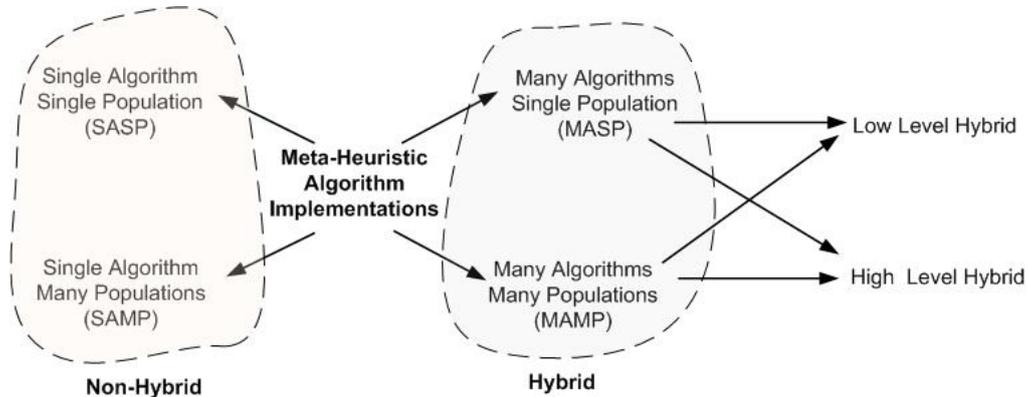

Figure 2. Population-based Meta-Heuristic Algorithm Implementations

Complementing the work by Talbi in [25], the two given grouping can be elaborated in further details as follows:

- MASP as Low-Level Hybrid (MASP-LLH)
  MASP-LLH implementation can be in the form of a relay (i.e., each meta-heuristic algorithm is sequentially applied) or cooperative (with no given ordering) applied to the same population. The advantage of MASP-LLH is that the user has direct control of the structure of whole combined algorithms. The limitation of MASP-LLH is that the developed hybrid can be too problem-specific.

  Examples of relay MASP-LLH include the work of Nasser et al. [26], Rambabu et al. [27], and Long et al. [28], respectively. The work of Nasser et al. [26] implement Flower Pollination Algorithm (FPA) [29] with the mutation operator. The work also adopts an elitism operator to improve the current poor solutions. Meanwhile, Rambabu et al. [27] integrate Artificial Bee Colony (ABC) [30] with Monarch Butterfly Optimization Algorithm (MBOA) [31]. Specifically, the MBOA is integrated as part of the employer bee adjusting phase preceding the onlooker bee phase. Long et al. [28] developed a hybrid algorithm by integrating Grey Wolf Optimizer (GWO) [32] with Cuckoo Search (CS) [33] for parameter extraction of solar photovoltaic models. In the work, the update of a new grey wolves' location is achieved via the CS levy flight operator.

  Meanwhile, examples of cooperative MASP-LLH are the work of Zamli et al. [23], Alotaibi [34], and Sharma et al. [35], respectively. Zamli et al. [23] develop a cooperative hybrid algorithm that allows low-level selection between the sine operator and the cosine operator from Sine Cosine Algorithm (SCA) [36], levy flight operator from Cuckoo Search Algorithm (CSA)[33] and crossover operator from Genetic Algorithm (GA) [37] using the Q-learning framework. Alotaibi [34] develops a low-level hybrid that integrates the Firefly Algorithm (FA) [38] with Jaya Algorithm (JA) [9] for video copyright protection. The selection of either FA or Jaya update is achieved based on the predefined trial constant. Initially, the algorithm uses FA whilst after reaching the trial constant, Jaya will take over. In other work, Sharma et al. [35] develop an ensemble of Butterfly Optimization Algorithm (BOA) [11] and Symbiosis Organisms Search (SOS) [39] where the global search ability of BOA and the local search ability of SOS are combined for solving the general global optimization problem.

- MASP as High-Level Hybrid (MASP-HLH)
  MASP-HLH implementation can also be in the form of a relay (i.e., each meta-heuristic algorithm is sequentially applied) or cooperative (with no given ordering) applied to the same population. MASP-HLH can often be associated with hyper-heuristic algorithms (termed *as (meta)-heuristic to choose (meta)-heuristic*) [40-43]. By using many (meta)-heuristic algorithms (or their associated search operators), the hyper-heuristic methodology can be considered as a form of hybridization. Focusing on the generalized methodology for solving optimization problems (i.e., owing to strict separation from implementation and problem domain), hyper-heuristic algorithms comes in two flavors namely generative hyper-heuristic algorithms and selective hyper-heuristic algorithms. The generative hyper-heuristic algorithm can customize its combination of new heuristic from a pool of possible heuristics. In contrast, the selective hyper-heuristic algorithms select the heuristics from a predefined set of heuristics. In essence, unlike selective hyper-heuristic algorithms, generative hyper-heuristic algorithms lend themselves to different type of optimization problems with minimal changes.

MASP-HLH grouping, however, is not exclusively mapped to only hyper-heuristic algorithms. The grouping can also include other forms of many algorithms' hybridization as long as each participating algorithm is treated as individual black-box components and with non-intersecting feature replacement of the host algorithm.

Examples of relay MASP-HLH implementation are the work of Luan et al. [44], Noori and Ghannadpour [45] and Lepagnot et al. [46], respectively. Luan et al. [44] hybridized the Genetic Algorithm (GA) [37] with Ant Colony Optimization (ACO) [47] to settle the supplier selection problem. This high-level hybrid improves the GA and ACO separately to enhance its efficiency and effectiveness. The work by Noori and Ghannadpour [45] adopt three levels of the high-level relay optimization process. The Genetic Algorithm (GA) [37] serves as the main optimization algorithm and Tabu Search (TS) [8] as an improvement method. In each level, heuristics incorporate local exploitation in the evolutionary search in order to solve the Multi-Depot Vehicle Routing Problem with Time Windows. In other work, Lepagnot et al. [46] proposes a high-level relay hybrid algorithm that combines the Multiple Local Search Algorithm (MLSA) [48] for dynamic optimization, the Success-History Based Adaptive Differential Evolution (SHADE) [49], and the Standard Particle Swarm Optimization (PSO) [50]. The hybrid algorithm is then subjected to a selected benchmark black-box optimization problem.

Concerning cooperative MASP-HLH, the work of Ahmad et al. in [41], and Zamli et al. in [51] and in [42] can be highlighted as relevant examples. Ahmad et al. [41] propose a Monte Carlo-based hyper-heuristic technique that embeds the Q-learning framework as an adaptive meta-heuristic selection and acceptance mechanism. The work adopts low-level search operations from the Cuckoo Search Algorithm (CSA) [33], Jaya Algorithm (JA) [9] and Flower Pollination Algorithm (FPA) [9]. In similar work, Zamli et al. [51] develop Tabu Search [8] based hyper-heuristic algorithm which rides on Teaching Learning-based Optimization Algorithm (TLBO) [52], CSA [33], and PSO [50]. As an extension of the work in [51], Zamli et al. [42] integrate the Mamdani fuzzy inference selection with its hyper-heuristic algorithm along with low-level search operations based on the FPA [9], JA [9], GA crossover [37] and TLBO [52].

- MAMP as Low-Level Hybrid (MAMP-LLH)
  MAMP-LLH lends itself toward parallel execution. Relay based MAMP-LLH is possible but often ignored as the approach does not promote parallelism. Cooperative based MAMP-LLH is preferable but there is potential overhead related to parallelism in terms of the need to coordinate and synchronize the contribution from each algorithm's population improvement. Being low-level, cooperative MAMP-LLH can also be too problem specific (e.g., integrating domain specific assumptions into the developed hybrid).

  Examples of cooperative MAMP-LLH include the work of Pourvaziri and Naderi [53], Zhou and Yao [54], and Chen et al. [55], respectively. Pourvaziri and Naderi [53] introduce a hybrid multi-population genetic algorithm for the dynamic facility layout problem, which adopts the local search mechanism from Simulated Annealing (SA) [4]. Zhou and Yao [54] develop a multi-population parallel self-adaptive differential Artificial Bee Colony (ABC) [30] algorithm where the distinct hybrid evolutionary operators borrowed from the Differential Evolution (DE) [56] are adopted during the evolution process. Meanwhile, Chen et al. [55] incorporate chaos strategy, multi-population, and DE [56] as part of low-level hybrid Harris Hawks Optimization (HHO) [57] implementation.

- MAMP as High-Level Hybrid (MAMP-HLH)

    MAMP-HLH also lends itself toward parallel execution but does not favor a relay-based implementation (similar to MAMP-LLH). Unlike MAMP-LLH, the interaction between meta-heuristic algorithms is often at a highlevel of abstraction (e.g. through black box parameter interface), hence, promoting better generalization. Similar parallelism issues need to be addressed in terms of the overhead of coordination and synchronization of the contribution from each algorithm's population improvement.

    Examples of cooperative MAMP-HLH include the work of Zhang et al. [22], Cruz-Chávez et al. [58], and Łapa et al. [59]. Zhang et al. [22] hybridized CSA [33] with DE [56] to solve constrained engineering problems which can find satisfactory global optima and avoid premature convergence. This work divides the population into two subgroups and adopts CSA and DE for these two subgroups independently. In another work, Cruz-Chávez et al. [58] present a hybrid GA [37] with collective communication using distributed processing for the job shop scheduling problem. In this hybrid, diversification is performed by iterative SA [4] and the intensification of the search space is made through genetic approximation. Meanwhile, Łapa et al. [59] propose a hybrid multi-population based approach where specified populations are processed using different population-based algorithms and synchronized accordingly for selecting the structure and parameters of the controller.

Summing up, all the aforementioned works suggest that hybridization is useful to enhance the search performance of the original meta-heuristic algorithm (i.e., in terms of balancing the exploration and exploitation of the search process). Taking the current work further, the next section highlights the HGSO implementation as the host algorithm for our hybridization.

## 3. Henry Gas Solubility Optimization Algorithm

HGSO is inspired by the solubility behavior of gases in liquids [60] based on the Henry's law [61]. This law states that "At a constant temperature, the amount of a given gas that dissolves in a given type and volume of liquid is directly proportional to the partial pressure of that gas in equilibrium with that liquid" [62]. Mathematically, Henry's law can be expressed by equation (1) whereby $S_g$ corresponds to the solubility of a gas:

$$S_g = H \times P_g \qquad (1)$$

where $H$ is Henry's constant, which is specific for the given gas-solvent combination at a given temperature, and $P_g$ represents the partial pressure of the gas.

Additionally, the relation between the Henry's constant and the temperature dependence of a system can be described with the Van't Hoff equation as follows:

$$\frac{d \ln H}{d(1/T)} = \frac{-\nabla_{sol} E}{R} \qquad (2)$$

where $\nabla_{sol} E$ is the enthalpy of dissolution, $R$ is the gas constant, and A and B are two parameters for T, which depends on H. Therefore, equation (1) can be simplified as (see Eq. (3)):

$$H(T) = \exp(B/T) \times A \qquad (3)$$

where H is a function of parameters A and B, which A and B are two parameters for T dependence of H. Alternatively, one can create an expression based on $H^\theta$ at the reference temperature T=298.15 K.

$$H(T) = H^\theta \times exp\left(\frac{-\nabla_{sol} E}{R}\left(\frac{1}{T} - \frac{1}{T^\theta}\right)\right) \qquad (4)$$

The Van't Hoff equation is valid when $\nabla_{sol}E$ is a constant, therefore, Eq. (4) can be reformulated as follows:

$$H(T) = exp\left(-C \times \left(\frac{1}{T} - \frac{1}{T^\theta}\right)\right) \times H^\theta \qquad (5)$$

Based on the said law, HGSO algorithm can be described in eight steps as follows:

***Step 1: Initialization process.*** The number of gases (population size $N$) and the positions of gases are initialized based on Eq. (6):

$$X_i(t+1) = X_{min} + r \times (X_{max} - X_{min}) \qquad (6)$$

where $X_{(i)}$ represents the position of the $i^{th}$ gas in population $N$, $r$ is a random number between 0 and 1, and $X_{max}, X_{min}$ are the upper bound and lower bound respectively of the problem, and $t$ is the current iteration number. The number of gas $i$, values of Henry's constant of type $j(H_j(t))$, partial pressure $P_{i,j}$ of gas $i$ in cluster $j$, and $\frac{\nabla_{sol}E}{R}$ constant value of type $j(C_i)$ are initialized using Eq. (7):

$$H_j(t) = l_1 \times rand(0,1), P_{i,j} = l_2 \times rand(0,1), C_j = l_3 \times rand(0,1) \qquad (7)$$

where, $l_1, l_2, l_3$ are defined as constants with values equal to (5E−02, 100, and 1E−02), respectively.

***Step 2: Clustering.*** The population agents are clustered equal to the number of gas types. Each cluster has similar gases and therefore has the same Henry's constant value($H_j$).

***Step 3: Evaluation.*** Each cluster $j$ is evaluated to identify the best gas and finally, the gases are ranked to obtain the optimal gas in the entire population.

***Step 4: Update Henry's coefficient.*** Henry's coefficient is updated according to Eq. (8) where $H_j$ is Henry's coefficient for cluster $j$, $T$ is the temperature, $T^\theta$ is a constant and equal to 298.15, and *iter* is the total number of iterations:

$$H_j(t+1) = H_j(t) \times exp\left(-C_j \times \left(\frac{1}{T(t)} - \frac{1}{T^\theta}\right)\right), T(t) = (-t/iter) \qquad (8)$$

***Step 5: Update solubility.*** The solubility is updated according to Eq. (9) where $S_{i,j}$ is the solubility of gas $i$ in cluster $j$ and $P_{i,j}$ is the partial pressure on gas $i$ in cluster $j$ and $K$ is a constant:

$$S_{i,j}(t) = K \times H_j(t+1) \times P_{i,j}(t) \qquad (9)$$

***Step 6: Update position.*** The position is updated using Eq. (10) where the position of gas $i$ in cluster $j$ is denoted as $X_{(i,j)}$, and $r$ and $t$ are a random constant and the iteration time, respectively:

$$X_{i,j}(t+1) = X_{i,j}(t) + F \times r \times \gamma \times \left(X_{i,best}(t) - X_{i,j}(t)\right)$$

$$+F \times r \times \alpha \times \left(S_{i,j}(t) \times X_{best}(t) - X_{i,j}(t)\right)$$

$$\gamma = \beta \times exp\left(-\frac{F_{best}(t)+\varepsilon}{F_{i,j}(t)+\varepsilon}\right), \varepsilon = 0.05$$

$$(10)$$

$X_{(i,best)}$ is the best gas $i$ in cluster $j$, whereas $X_{best}$ is the best gas in the swarm. These two parameters are responsible for balancing exploration and exploitation abilities. Additionally, $\gamma$ is the ability of gas $i$ in cluster $j$ to interact with the gases in its cluster, $\alpha$ is the influence of other gases on gas $i$ in cluster $j$ and equal to 1 and $\beta$ is a constant. $F_{(i,j)}$ is the fitness of gas $i$ in cluster $j$, in contrast $F_{best}$ is the fitness of the best gas in the entire population. $F$ is the flag that changes the direction of the search agent and provides diversity.

***Step 7: Escape from local optimum.*** This step ranks and selects the number of worst agents ($N_w$) using Eq. (11) to escape from local optimum where $N$ is the number of search agents:

$$N_w = N \times (rand(c_2 - c_1) + c_1), \quad c_1 = 0.1, \quad c_2 = 0.2 \tag{11}$$

***Step 8: Update the position of the worst agents.*** The position update is given by Eq. (12).

$$X_{(i,j)} = X_{\min(i,j)} + r \times (X_{\max(i,j)} - X_{min(i,j)}) \tag{12}$$

where, $X_{(i,j)}$ is the position of gas $i$ in cluster $j$, $r$ is a random number and $X_{\min}$, $X_{\max}$ are the lower bound and upper bound of the problem, respectively.
The pseudocode of HGSO is summarized in Figure 3.

---

[1]. **begin**
[2].    Initialize population $X_i(i = 1, 2, \ldots, N)$, number of gas types $i, H_j, P_{i,j}, C_j, l_1, l_2$ and $l_3$.
[3].    Divide the population agents into a number of gas types (cluster) with the same Henry's constant value ($H_j$)
[4].    Evaluate each cluster $j$.
[5].    Get the best gas $X_{i,best}$ in each cluster, and the best search agent $X_{best}$.
[6].    **while** (stopping criteria not met (i.e. $t < Max_{iteration}$))
[7].       **for** each search agent **do**
[8].          Update the position of individual search agent using Eq. (10).
[9].       **end for**
[10].       Update Henry's coefficient of each gas type using Eq. (8).
[11].       Update solubility of each gas using Eq. (9).
[12].       Rank and select the number of worst agents using Eq. (11).
[13].       Update the position of the worst agents using Eq. (12).
[14].       Update the best gas $X_{i,best}$, and the best search agent $X_{best}$
[15].       $t = t + 1$
[16].    **end while**
[17].    **return** $X_{best}$
[18]. **end**

Figure 3. HGSO Pseudocode

## 4. The Proposed Hybrid HGSO

The proposed Hybrid HGSO (HHGSO) can be visualized as in Figure 4. At any iteration, HHGSO provides dynamic meta-heuristics-to-cluster-mapping. For instance, as iteration=0, the meta-heuristic-to-cluster-mapping could be different from the one at iteration=J or iteration=K. The dynamic meta-heuristic-to-cluster-mapping is achieved through penalized and reward model with adaptive probability.

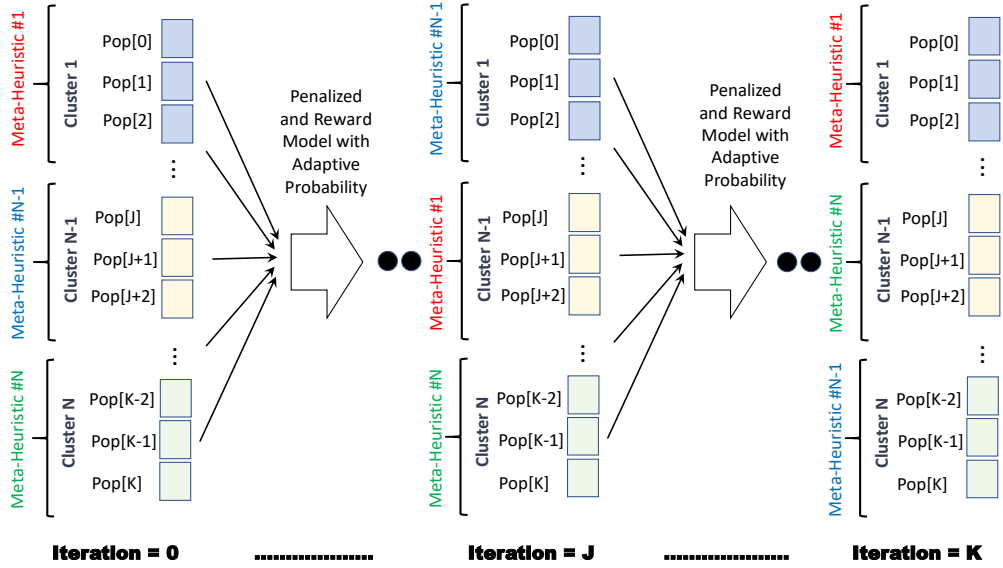

Figure 4. Dynamic Meta-Heuristic-to-Cluster-Mapping

### 4.1. Penalized and Reward Model with Adaptive Probability

In a nut shell, the penalized and reward model considers the co-operative performance of a particular meta-heuristic-to-cluster mapping as a whole. If the co-operative performance at iteration = K realizes an improvement of the population's global best, then the meta-heuristic-cluster-mapping will be rewarded and be maintained for the next iteration = K+1. Otherwise, the meta-heuristic-to-cluster-mapping will be updated randomly based on adaptive probability *p* as follows (see Eq. (13):

$$p = p_{max} + \frac{t(p_{min} - p_{max})}{Max_{iteration}} \tag{13}$$

where the minimum probability $p_{min} = 0.2$ and the maximum probability $p_{max} = 0.8$. *t* is the current iteration and $Max_{iteration}$ is the maximum iteration.

In the early part of the search iteration, the probability *p* will be large meaning that the meta-heuristic-to-cluster-mapping is likely to change whenever the performance of the overall search is poor. In this manner, each cluster can have a different update operator from more than one meta-heuristic algorithm to ensure sufficient exploration. Towards the end during convergence, the probability *p* becomes small minimizing any significant change to the meta-heuristic-to-cluster-mapping supporting better exploitation.

### 4.2. Constituent Meta-Heuristic Algorithms

Apart from HGSO itself as host, HHGSO leverage on four other meta-heuristic algorithms as its hybridization vehicle (i.e., Jaya, Sooty Tern Optimization Algorithm (STOA), Butterfly Optimization Algorithm (BOA) and Owl Search Algorithm (OSA)). The brief description of each meta-heuristic algorithm is presented in the following sub-sections.

#### 4.2.1. Jaya Algorithm (JA)

Jaya is parameter free meta-heuristic algorithm developed by Rao [9]. Jaya works by establishing the solution to problems through avoiding the worst solutions and moving towards the best optimal solution. Jaya modifies its solutions based on the best and worst solutions using Eq. (14).

$$X_i(t+1) = X_i(t) + r_1(X_{best} - |X_i(t)|) - r_2(X_{worst} - |X_i(t)|) \tag{14}$$

where $X_{best}$ is the value of the variable for $F_{best}$ and $X_{worst}$ is the value of the variable for $F_{worst}$ and $X_i(t+1)$ as the updated $i^{th}$ value of $X_i(t)$. Here, $r_1$ and $r_2$ are the two random scaling factors in the range [0,1].

### 4.2.2. Sooty Tern Optimization Algorithm (STOA)

STOA is the population based meta-heuristic algorithm, proposed by Dhiman and Kaur [10]. The main idea behind STOA is to mimics the migration and attacking behaviors of sea bird sooty tern in nature. The migration behavior (represented in Eq. (15) until Eq. (19)) and attacking behavior (represented in Eq. (20) until Eq. (24)) of sooty terns in STOA are mathematically simulated as follows:

$$C_i = S_A \times X_i(t) \tag{15}$$

$$S_A = C_f - (t \times (C_f/Max_{iteration})) \tag{16}$$

where $C_i$ is the position of search agent, $X_i(t)$ represents the current position of search agent at iteration $t$, $S_A$ indicates the movement of the search agent in a given search space. $C_f$ is a controlling variable to adjust the $S_A$ which is linearly decreased from $C_f$ to 0. $t$ is the current iteration ($t = 0,1,\ldots Max_{iteration}$) and $Max_{iteration}$ is the maximum number of iterations.

$$M_i = C_B \times (X_{best}(t) - X_i(t)) \tag{17}$$

$$C_B = 0.5 \times rand(0,1) \tag{18}$$

where $M_i$ represents the different location of the search agent $X_i(t)$ towards the best fittest search agent $X_{best}(t)$. $C_B$ is a uniformly distributed random variable (responsible for better exploration).

$$D_i = C_i \times M_i \tag{19}$$

where $D_i$ defines the gap between the search agent and best fittest search agent

$$x' = R_{adius} \times \sin(m) \tag{20}$$
$$y' = R_{adius} \times \cos(m) \tag{21}$$
$$z' = R_{adius} \times m \tag{22}$$
$$r = u \times e^{kv} \tag{23}$$

where $R_{adius}$ represents the radius of each turn of the spiral, $m$ represents the variable lies between the range of $[0 \leq k \leq 2\pi]$. $u$ and $v$ are constants to define the spiral shape, and $e$ is the base of the natural logarithm.

The candidate solution update is given in Eq. (24) as follows:

$$X_i(t) = (D_i(t) \times (x' + y' + z')) \times X_{best}(t) \tag{24}$$

where $X_{best}(t)$ is the best fittest search agent. $D_i(t)$ defines the gap between the search agent and best fittest search agent. $x'$, $y'$, and $z'$ represent the angle of attack.

### 4.2.3. Butterfly Optimization Algorithm (BOA)

Butterfly Optimization Algorithm (BOA) is a population-based meta-heuristic algorithm proposed by Arora et al. [11]. The BOA mimics the foraging and the social behavior of the butterflies. In a nut shell, butterflies use their senses for finding foods, searching for a mating partner, migrating from one place to another, and escaping from enemies.

BOA exploits the fragrance as a function of the physical intensity stimulus as part of its movement based on Eq. (25).

$$f_i = cI^a \tag{25}$$

where $c = 0.01$ is a sensory modality, $I$ is the stimulus intensity typically upper-lower bound, $a$ is power exponent linearly updated from 0.1 to 0.2.

The global candidate solution update is given by Eq. (26).

$$X_i(t+1) = X_i(t) + (r^2 \times X_{best} - X_i(t)) \times f_i \tag{26}$$

where $X_i(t)$ is the solution vector $x_i$ for $i^{th}$ butterfly in iteration $t$. $X_{best}$ represents the current best solution found among all the solutions in the current iteration. $f_i$ represents the fragrance of $i^{th}$ butterfly, and r is a random number between [0, 1].

Complementing the global candidate solution update, BOA as defines local candidate update as in Eq. (27).

$$X_i(t+1) = X_i(t) + (r^2 \times X_j(t) - X_k(t)) \times f_i \tag{27}$$

where $X_j(t)$ and $X_k(t)$ are $j^{th}$ and $k^{th}$ butterflies from the solution space.

### 4.2.4. Owl Search Algorithm (OSA)

Owl Search Algorithm (OSA) is a population-based meta-heuristic algorithm developed by Jain et al [12]. OSA mimics the hunting mechanism of the owls in the dark. Like other population-based meta-heuristics, OSA stores the initial positions and fitness values of all owls in a two-dimensional matrix (i.e., $X$ and $F$). The size of each matrix is $n \times d$ where n is the number of owls and $d$ represents the dimension of search space. The fitness value of the owls position directly relates to the sound intensity. The intensity of $i^{th}$ owl can be calculated by Eq. (28).

$$I_i = \frac{F_i - w}{b - w} \tag{28}$$

where $F_i$ is the fitness of $i^{th}$ owl, $w$ minimum fitness, and $b$ represents the maximum fitness.

Eq. (29) represents the distance information of each owl and prey. The change in intensity of the $i^{th}$ owl can be computed using Eq. (30).

$$R_i = \|X_i, V\|_2 \tag{29}$$

$$Ic_i = \frac{I_i}{R_i^2} + rand(0,1) \tag{30}$$

where $V$ is the location of prey calculated from fittest owl, $R_i$ is the distance of pray from the owl $X_i$.

The owls candidate solution update is given in Eq. (31).

$$X_i(t+1) = \begin{cases} X_i(t) + \beta \times Ic_i \times |\alpha V - X_i(t)|, if\ p_{vm} < 0.5 \\ X_i(t) - \beta \times Ic_i \times |\alpha V - X_i(t)|, if\ p_{vm} < 0.5 \end{cases} \tag{31}$$

where $p_{vm}$ is the probability of vole movement, $\alpha$ is a uniformly distributed random number in the range [0, 0.5], and $\beta$ is a linearly decreasing constant from 1.9 to 0. $Ic_i$ is the change in intensity for $i^{th}$ owl. $V$ is the location of prey, which is achieved by the fittest owl (i.e., $X_{best}$).

## 4.3. HHGSO Implementation

Considering the penalized and reward model with adaptive probability as well as all the constituent meta-heuristic algorithms, HHGSO pseudocode can be summarized in Figure 5.

Referring to Figure 5, the HHGSO algorithm starts with defining the algorithm list, cluster *N_size*, each meta-heuristic algorithm's parameter initialization along with population initialization (see line 1 until line 5). The main iteration loop starts in line 6. The algorithm-to-cluster-mapping is tracked by the division index variable (see line 7). Modulo division is used to ensure that the division index is mapped correctly to the corresponding algorithm in use at any particular iteration. The current running algorithm is mapped to the algorithm list as determined by division index variable (see line 10 and line 13). The selection of the running meta-heuristic algorithm follows accordingly (refer to line 15 until 34). To go out of local optima, HHGSO will update $N_x$ poor solution with the new random ones (see line 36). As each cluster maintains its own local cluster best $X_{cluster\ best}$ and the selection of the global best $X_{best}$ will be made amongst the local cluster best (see line 37). If the overall cooperative performance improves the same meta-heuristic-to-cluster is maintained. Otherwise, with decreasing adaptive probability, the ordering of algorithm list will be reshuffled in order to generate a new meta-heuristic-to-cluster-mapping (line 40 until line 43). If there are more clusters than the list of defined algorithms, leftover clusters will be mapped to HGSO as the host algorithm (line 42). The iteration will stop when $Max_{iteration}$ or maximum fitness evaluation is reached (line 46). In the end, the global best result $X_{best}$ will be returned upon completion (refer to line 48).

```
[1].  begin
[2].     define algorithm list = #"Jaya", "SootyTern", "Owl", "Butterfly", "HenryGas"#
[3].     define cluster N_size (default = 5)
[4].     initialize all algorithms' defined parameters accordingly
[5].     initialize population $X_i (i = 1, 2, \ldots, Max_{pop})$
[6].     while (stopping criteria not met (i.e. $t < Max_{iteration}$))
[7].        set division_idx=1;
[8].        for each search agent (i.e. $i = 1, 2, \ldots, Max_{pop}$) do
[9].           if (the first agent)
[10].             current_running_algorithm=algorithm_list[division_idx]
[11].          else if (i % N_size==0)
[12].             division_idx++
[13].             current_running_algorithm=algorithm_list[division_idx]
[14].          end if
[15].          if (current_running_algorithm ="Jaya")
[16].             generate new candidate solution using Jaya based on Eq. (14)
[17].             maintain cluster best $X_{cluster\ best}$
[18].          else if (current_running_algorithm="SootyTern")
[19].             update Sooty Tern parameters based on Eq. (15) until Eq. (23)
[20].             generate new candidate solution using Sooty Tern based on Eq. (24)
[21].             maintain cluster best $X_{cluster\ best}$
[22].          else if (current_running_algorithm ="Butterfly")
[23].             update Butterfly parameters based on Eq. (25)
[24].             generate new candidate solution using Butterfly based on Eq. (26) or Eq. (27)
[25].             maintain cluster best $X_{cluster\ best}$
[26].          else if (current_running_algorithm="Owl")
[27].             update Owl parameters based on Eq. (28) until Eq. (30)
[28].             generate new candidate solution using Owl based on Eq. (31)
[29].             maintain cluster best $X_{cluster\ best}$
[30].          else if (current_running_algorithm="HenryGas")
[31].             update Henry Gas parameters based on Eq. (7) until Eq. (10)
[32].             generate new candidate solution using Henry Gas based on Eq. (6)
[33].             maintain cluster best $X_{cluster\ best}$
[34].          end if
[35].       end for
[36].       select $N_w$ =worst population based on Eq. (11) and update them using Eq. (12)
[37].       select one $X_{best}$ from all $X_{cluster\ best}$
[38].       if (non-improving cooperation $X_{global\_best\_new}$ is not better than $X_{global\_best\_old}$)
[39].          update adaptive threshold probability based on Eq. (13)
[40].          if (random [0,1] <threshold probability)
[41].             reshuflle ordering of algorithm list
[42].             all left over clusters assigned to HenryGas (i.e. N_size>length of algorithm list)
[43].          end if
[44].       end if
[45].       $t=t+1$
[46].       break while loop when (fitness evaluation>=max_fitness_evaluation i.e. $Max_{fit\ eval}$)
[47].    end while
[48].    return the global best solution $X_{best}$
[49]. end
```

Figure 5. Hybrid HGSO Pseudocode

## 5. Empirical Evaluation

We have subjected our work under intensive evaluation. Our goals of the evaluation experiments are threefold: (1) to characterize the performance HHGSO against the original HGSO algorithm and its constituent meta-heuristic algorithms based on the benchmark team formation problem; (2) to benchmark HHGSO against hyper-heuristic algorithms (as cooperative MASP-HLH implementation) based on the benchmark combinatorial t-way test data generation; (3) to assess the effect of cluster size on the meta-heuristic-algorithm-to-cluster-mapping of HHGSO.

In line with the aforementioned goals, we focus on answering the following research questions:

- RQ1: How is the performance of HHGSO compared to that of the original HGSO and its participating constituent algorithms?
- RQ2: What is the effect of cluster size with the meta-heuristic-algorithm-to-cluster-mapping?
- RQ3: How good is the performance of HHGSO against its cooperative MASP-HLH counter parts (i.e., hyper-heuristic algorithms)?
- RQ4: Is there any overhead in terms time performance penalty of HHGSO implementation as compared to its constituent algorithms?
- RQ5: How generalized can HHGSO implementation be for solving general optimization problems?

### 5.1. Experimental Benchmark Setup

We adopt an environment consisting of a machine running Windows 10, with a 2.9 GHz Intel Core i5 CPU, 16 GB 1867 MHz DDR3 RAM, and 512 GB flash storage throughout all our experiments. We implement our HHGSO in the Java programming language.

To ensure fairness, we have adopted different settings on population size, maximum iteration as well as maximum fitness evaluation for relevant experiments and their related research questions. For RQ1, we have adopted the population size $N = 50$ with the maximum iteration $Max_{iter} = 100$. Here, we limit the maximum fitness evaluation $Max_{fit\ eval}=2500$.

Meanwhile, for RQ3, we have adopted the population size $N = 20$ with the maximum iteration $Max_{iter} = 100$ to ensure the same setting as the original benchmark experiments. In this case, the maximum fitness evaluation $Max_{fit\ eval}=2000$.

Apart from the population size and maximum iteration, other algorithm-specific parameter settings of all constituent meta-heuristic algorithms are summarized in Table 1.

Table 1. Parameter settings values for algorithms

| Algorithms | Parameters | Value |
|---|---|---|
| Jaya Algorithm (JA) | No algorithm-specific parameters | - |
| Sooty Tern Optimization Algorithm (STOA) | Controlling Variable ($C_f$) | [2, 0] |
| | Random Variable ($C_B$) | [0, 0.5] |
| | Constants $u$ and ($v$) | 1 |
| | Variable ($k$) | [0, 2π] |
| Butterfly Optimization Algorithm (BOA) | Sensory Modality ($c$) | [0, 1] |
| | Power Exponent ($a$) | [0, 1] |
| | Switch Probability ($p$) | 0.8 |
| Owl Search Algorithm (OSA) | Probability of Vole Movement ($p_{vm}$) | [0, 1] |
| | Uniformly Distributed Random Number ($\alpha$) | [0, 0.5] |
| | Linearly Decreasing Constant ($\beta$) | [1.9, 0] |
| Henry Gas Solubility Optimization Algorithm (HGSO) | Cluster | 5 |
| | Constant ($l_1$) | 5E-02 |
| | Constant ($l_2$) | 100 |
| | Constant ($l_3$) | 1E-02 |
| | Constant ($T^\theta$) | 298.15 |
| | Constant ($K$) | 1.0 |
| | Influence of Gas ($\alpha$) | 1.0 |
| | Constant ($\beta$) | 1.0 |
| | Constant ($C_1$) | 0.1 |
| | Constant ($C_2$) | 0.2 |

For statistical significance, we have executed HHGSO and all its constituent 30 times and reported the best and best worst results as well as best average using these runs (as bold cells). Whenever possible, we also report the best average execution time (also as bold cells). For RQ1, we also reported the number of team members.

### 5.2. Case Study Objects Selection and Experimental Procedure

Our case study objects relate to two Search-based Software Engineering problem namely the team formation problem and the combinatorial t-way test suite generation. The discussion on RQ1 till RQ3 will be based on the results of the experiments while the discussion on RQ4 and RQ5 will be based on the lessons learned from undertaking the work.

### 5.2.1. Team Formation Problem

The team formation problem can be seen as a set covering problem (SCP). Considered NP hard problem, the mathematical formulation of the set covering problem is as follows.

Let a universe of elements $E = \{e_1, ..., e_m\}$ and let the collection of subset $S = \{s_1, ..., s_n\}$ where $s_j \subseteq E$ and $\bigcup s_j = E$. Each set $s_j$ covers at least one element of $E$ and has an associated cost $c_j > 0$. The objective is to find a sub-collection of sets $X \subseteq E$ that covers all of the elements in $E$ at a minimal cost.

Let $A^{m \times n}$ be a zero-one matrix where $a_{ij} = 1$ if element $i$ is covered by set $j$ and $a_{ij} = 0$ otherwise. Let $X = \{x_1, ..., x_n\}$ where $x_j = 1$ if set $s_j$ (with cost $c_j > 0$) is part of the solution and $x_j = 0$ otherwise.

$$\text{Minimize } \sum_{j=1}^{n} c_j x_j \quad (32)$$

Subject to
$$1 \leq \sum_{j=1}^{n} a_{ij} x_j, i = 1, ..., m \quad (33)$$
$$x_j \in \{0,1\} \quad (34)$$

In the team formation problem, the goal is to form a team that covers all the required skills from the given search space of individual experts with certain defined skills. Based on the model from Lappas et al. [63], the costs of interaction between two experts (A and B) can be calculated using Eq. (35).

$$\text{Interaction Cost between Expert's A and B} = 1 - \frac{\text{Skills of A} \cap \text{Skills of B}}{\text{Skills of A} \cup \text{Skills of B}} \quad (35)$$

The best team is the one with the most minimum interaction costs between experts in the team. For RQ1 and RQ2, we subject HHGSO to two benchmark team formation problem data set involving IMDB[64] and DBLP [65]. The IMDB data set is a database of the movie actors and their roles by genre owned by Amazon. The cleaned data set consists of 1014 names of actors and unique 28 roles by genre. Meanwhile, the DBLP data set is a bibliographic database of scientific publications. The cleaned data set includes 5641 authors' information with 3887 unique skills.

For our evaluation, we have adopted three sets of skills to look for. For IMDB, we adopt the 8, 16, and 24 skills set. Meanwhile, for DBLP, we have adopted the 30, 60, and 90 skills set. To ensure fairness of comparison, apart from running on the same platform, we have implemented the JA, STOA, BOA, OA and HGSO using the Java programming language with the same data structure as HHGSO. For this reason, we are also able to report the time performance in addition to costs.

### 5.2.2. Combinatorial Test Suite Generation

Considered NP hard, the combinatorial test suite generation is an optimization problem with the aim of generating the most minimum $t$-way test interaction test size. In any test suite, every $t$-way interaction combinations must be covered at least once. The rationale behind $t$-way testing is that not every parameter contributes to faulty conditions, and many faulty condition can be exposed by considering the interaction of only a few parameters.

The mathematical formulation fo the $t$-way test generation problem can be expressed as in Eq. (36).

$$\text{Minimize} \quad f(Z) = |\{I \text{ in } VIL: Z \text{ covers } I\}| \quad (36)$$

Subject to

$$Z = Z_1, Z_2, \ldots Z_i \text{ in } P_1, P_2, \ldots \ldots P_i; \ i = 1, 2, \ldots N$$

where, $f(Z)$ is an objective functions (or the fitness evaluation ), $Z$ (i.e., the test case candidate) is the set of decision variables $Z_i$, $VIL$ is the set of non-covered interaction tuples ($I$), the vertical bars $|\cdot|$ represent the cardinality of the set and the objective value is the number of non-covered interaction tuples covered by $Z$, $P_i$ is the set of possible range of values for each decision variable, that is, $P_i$ = discrete decision variables ($Z_i(1) < Z_i(2) < \ldots \ldots < Z_i(K)$); N is the number of decision variables (i.e., parameters); and K is the number of possible values for the discrete variables.

Concerning notation, the $t$-way test suite generation is often expressed in term of Covering Array (CA) notations. The notation CA has four main parameters, namely, $S$, $t$, $p$, and $v$ (i.e., CA $(S, t, v^P)$. CA is a matrix of size $S \times P$. Here, the symbols $t$ refers to the interaction strength, $S$ represents the test cases (rows), $P$ is known as number of parameters (columns) and $v$ refers to the number of $CA$ values for a specific $P$. For example, CA $(S, 2, 3^4)$ can be seen as $S \times 4$ array that covers the test suite. In this case, the test suite covers $t = 2$ with three $v$ values and four $p$ parameters, $S = 3 \times 3 = 9$ test cases.

For RQ3, we have subjected HHGSO to the following benchmark experiments (from reference [42]) as follows:
- A set of $CA_1$ (N; 2, $3^{13}$), $CA_2$ (N; 2, $10^{10}$), $CA_3$ (N; 3, $3^6$), $CA_4$ (N; 3, $6^6$), $CA_5$ (N; 3, $10^6$), $CA_6$ (N; 3, $5^2 4^2 3^2$)
- CA (N; 2, $3^k$) where $k$ is varied from 3 to 12
- CA (N; 3, $3^k$) where $k$ is varied from 4 to 12
- CA (N; 4, $3^k$) where $k$ is varied from 5 to 12
- CA (N; 2, $v^7$) where $v$ is varied from 2 to 7
- CA (N; 3, $v^7$) where $v$ is varied from 2 to 7
- CA (N; 4, $v^7$) where $v$ is varied from 2 to 7

The size performances for cooperative MASP-LLH algorithms (i.e., Exponential Monte Carlo hyper-heuristic with Counter, Modified Choice Function hyper-heuristic, Improvement Selection Rules hyper-heuristic, and Fuzzy Inference Selection hyper-heuristic) are taken directly from the original reference [42]. As no time performances are reported in the original reference, we also do not report our time performance for HHGSO. Unlike time performance (which depends on implementation language, data structure, system configuration and the running environment), size performance is absolute. For this reason, the size performance can also give meaningful indication of HHGSO performance against its cooperative MASP-LLH counter parts.

## 6. Results

Our results will be aligned to the given research questions as follows:
- **RQ1: How is the performance of HHGSO compared to that of the original HGSO and its participating constituent algorithms?**

Referring to Table 2 and 3, HHGSO outperforms HGSO and all its constituent algorithms. To be specific, HHGSO obtain the best costs for all cases with the exception of DBLP with 30 required skills (see Table 3). Here, HGSO has the best costs (i.e., 1464.35) although having the same number of team members (i.e., 55) with HHGSO. In terms of average costs, HHGSO outperforms all the other algorithms. Unlike best costs which can be influenced by chance, average costs show the consistent performance of HHGSO as compared to other algorithms.

Putting HHGSO aside, HGSO comes in as the runner up. The fact that HHGSO and HGSO perform better than its constituent algorithms can be attributed to the fact that there are potentially more diversity in the population of solutions. HHGSO, in particular, enjoys five different candidate update operators (with different displacements) from Jaya, Sooty Tern Algorithm, Butterfly Optimization Algorithm, Owl Search Algorithm and HGSO. The probabilistic changes in the update operators (owing to dynamic-meta-heuristic-to-cluster mapping) ensure that the search process can easily go out-of-local optima.

The performance of Jaya, Sooty Tern Algorithm, and Owl Search Algorithm is at par with each other as they have mixed results as far as the average costs is concerned. Butterfly Optimization Algorithm performs the worst as its average costs is no better than any of the compared algorithms. On a positive note, the Butterfly Optimization Algorithm manages to outperform other algorithms in terms of the average execution time for the case of DBLP with 90 required skills. Overall, Owl Search Algorithm has the best average execution time.

Table 2. Comparative Performance of HGSO and its Constituent Algorithms based on IMDB Data Set for 8, 16, and 24 required skills

| | JA | STOA | BOA | OSA | HGSO (cluster size = 5) | HHGSO (cluster size = 5) |
|---|---|---|---|---|---|---|
| **8 required skills** | | | | | | |
| **Metrics Measurement** | | | | | | |
| Best Cost | 4.32 | 4.04 | 4.26 | 4.37 | 3.81 | **2.49** |
| No of Team Members | 4 | 4 | 4 | 4 | 4 | **3** |
| Ave Cost | 7.08 | 6.79 | 7.29 | 6.77 | 6.60 | **6.58** |
| Ave Time(sec) | 3.25 | 3.05 | 3.20 | 3.03 | **3.00** | 3.01 |
| **16 required skills** | | | | | | |
| **Metrics Measurement** | | | | | | |
| Best Cost | 13.98 | 13.89 | 20.65 | 15.97 | 12.96 | **11.60** |
| No of Team Members | 7 | **7** | 8 | 7 | **6** | 6 |
| Ave Cost | 28.67 | 27.87 | 33.59 | 28.26 | 27.28 | **26.73** |
| Ave Time (sec) | 3.20 | 3.30 | 3.24 | 3.25 | **2.99** | 3.03 |
| **24 required skills** | | | | | | |
| **Metrics Measurement** | | | | | | |
| Best Cost | 17.71 | 24.22 | 32.11 | 28.40 | 23.76 | **22.80** |
| No of Team Members | 7 | 8 | 9 | 9 | **8** | 8 |
| Ave Cost | 46.45 | 44.69 | 54.93 | 47.50 | 43.79 | **43.61** |
| Ave Time (sec) | 3.58 | 3.26 | 3.34 | 3.31 | 3.05 | **2.96** |

Table 3. Comparative Performance of HGSO and its Constituent Algorithms based on DBLP Data Set for 30, 60, and 90 required skills

| | JA | STOA | BOA | OSA | HGSO (cluster size = 5) | HHGSO (cluster size = 5) |
|---|---|---|---|---|---|---|
| **30 required skills** | | | | | | |
| **Metrics Measurement** | | | | | | |
| Best Cost | 342.64 | 319.63 | 368.68 | 343.43 | 318.93 | **317.16** |
| No of Team Members | 27 | **26** | 28 | 27 | **26** | 26 |
| Ave Cost | 360.28 | 345.34 | 369.08 | 352.09 | 352.39 | **340.10** |
| Ave Time (sec) | 487.72 | 427.30 | 487.06 | **391.48** | 455.82 | 479.87 |
| **60 required skills** | | | | | | |
| **Metrics Measurement** | | | | | | |
| Best Cost | 1464.86 | 1515.32 | 1522.00 | 1516.40 | **1464.35** | 1469.29 |
| No of Team Members | **55** | 56 | 56 | 56 | **55** | 55 |
| Ave Cost | 1537.14 | 1537.39 | 1574.25 | 1535.73 | 1520.11 | **1502.70** |
| Ave Time (sec) | 495.02 | 505.35 | **463.22** | 480.65 | 479.35 | 472.26 |
| **90 required skills** | | | | | | |
| **Metrics Measurement** | | | | | | |
| Best Cost | 3273.96 | 3195.21 | 3524.01 | 3268.25 | 3266.92 | **3035.01** |
| No of Team Members | 82 | 81 | 85 | 82 | 82 | **79** |
| Ave Cost | 3411.86 | 3330.80 | 3578.53 | 3406.82 | 3404.61 | **3327.03** |
| Ave Time (sec) | 444.81 | 509.31 | 472.02 | **406.44** | 504.20 | 421.22 |

- **RQ2: What is the effect of cluster size with the meta-heuristic-algorithm-to-cluster-mapping?**

To answer RQ2, there is a need to deliberate on two main issues. The first issue relates to the effect of cluster size on the HHGSO overall performance of the search process  The second issue relates to the effect of cluster size to the algorithm mapping (e.g., whether or not there is a certain preference on a particular constituent algorithm).

Concerning the first issue, the metrics measurements for cluster size=1 until 6 from Table 4 for IMDB and Table 5 for DBLP provides some indication on the effect of cluster size to the HHGSO overall performance. Specifically, we are interested on the performance of HHGSO with cluster size=5. There reason is that there is a one-to-one mapping of each participating meta-heuristic algorithm with the defined cluster.

From the given results, HHGSO with cluster size=5 outperforms HHGSO with other cluster size as far as the average cost is concerned.  There is only one instance where HHGSO does not have the best average costs, that is, involving IMDB with 8 required skills (see Table 3). In this case, HHGSO with cluster size=4 has outperformed HHGSO with cluster size=5. We consider this as outlier as HHGSO with cluster size = 4, is not performing well in other instances involving other given IMDB or DBLP datasets with other defined skills to find. The same observation can be seen as far as the best cost is concerned. HHGSO with cluster size=5 give the best cost for all cases with the exception of  one case involving DBLP with 60 required skills (see Table 4). Again, the fact that HHGSO with cluster=3 gives the best costs can also due to outlier as it does not perform well in other instances.

We conclude that as far as HHGSO with cluster size=1 till 4 and 6 is concerned, there is no evidence of better performance than that of HGGSO with cluster size=5 apart from having better execution times. When the cluster size definition is less than the participating algorithms, the algorithm-to-cluster-mapping becomes too randomized resulting into some algorithms be selected multiple times even though they may not be the performing ones (as side effect of the dynamic probability). For this reason, the performance of HHGSO with cluster size < 5 is often poorer as compared to HHGSO with cluster size =5.  Having HHGSO variants with cluster size > participating algorithms also appear counter-productive as the extra cluster will be biased toward HHGSO update operators. This is reflected by the results tabulated in Table 4 and 5, respectively.

It is interesting to note that in many part of the results, the number of team members is the same for almost all cluster sizes. For example, in the case of IMDB with 8 required skills, HHGSO with cluster size-4, 5 and 6 has the same number of team members of 3 yet with different costs. The same observation can be seen for DBLP cases also. As the skills are not unique, different combination of team is possible although at different costs.

Table 4. Effect of Cluster Size based on IMDB Data Set for 8, 16, and 24 required skills

| | HHGSO | | | | | |
|---|---|---|---|---|---|---|
| | Cluster Size=1 | Cluster Size=2 | Cluster Size=3 | Cluster Size=4 | Cluster Size=5 | Cluster Size=6 |
| **8 required skills** | | | | | | |
| **Metrics Measurement** | | | | | | |
| Best Cost | 4.17 | 4.10 | 3.95 | 2.66 | **2.49** | **2.49** |
| No of Team Members | 4 | 4 | 4 | **3** | **3** | **3** |
| Ave Cost | 6.64 | 6.63 | 6.32 | **6.48** | 6.58 | 6.59 |
| Ave Time (sec) | **2.89** | 3.41 | 2.92 | 3.15 | 3.01 | 2.95 |
| **Ave Execution Distribution** | | | | | | |
| HGSO | 20.00% | 12.07% | 21.96% | 21.17% | 20.37% | 32.99% |
| JA | 30.00% | 26.78% | 15.17% | 20.65% | 19.60% | 16.67% |
| STOA | 26.67% | 21.84% | 21.26% | 17.42% | 20.04% | 16.78% |
| BOA | 10.00% | 16.44% | 27.01% | 21.51% | 19.67% | 16.55% |
| OSA | 13.33% | 22.87% | 14.60% | 19.25% | 20.32% | 17.01% |
| **16 required skills** | | | | | | |
| **Metrics Measurement** | | | | | | |
| Best Cost | 15.87 | 12.39 | 11.87 | 15.02 | **11.60** | **11.60** |
| No of Team Members | 7 | **6** | **6** | 7 | **6** | **6** |
| Ave Cost | 28.11 | 27.97 | 27.90 | 27.99 | **26.73** | 26.90 |
| Ave Time (sec) | 3.00 | **2.88** | 3.00 | 2.99 | 3.03 | 2.91 |
| **Ave Execution Distribution** | | | | | | |
| HGSO | 50.00% | 26.44% | 40.23% | 19.62% | 20.46% | 33.67% |
| JA | 10.00% | 26.44% | 15.40% | 19.88% | 19.44% | 16.21% |
| STOA | 10.00% | 11.49% | 17.93% | 19.88% | 20.33% | 16.55% |
| BOA | 16.67% | 8.39% | 10.92% | 21.36% | 19.47% | 16.44% |
| OSA | 13.33% | 27.24% | 15.52% | 19.26% | 20.30% | 17.13% |
| **24 required skills** | | | | | | |
| **Metrics Measurement** | | | | | | |
| Best Cost | 24.28 | 24.69 | 24.42 | 22.92 | **22.80** | 28.14 |
| No of Team Members | **8** | **8** | **8** | **8** | **8** | 9 |
| Ave Cost | 48.09 | 47.16 | 45.75 | 45.94 | **43.61** | 45.89 |
| Ave Time (sec) | 3.11 | 3.21 | 2.97 | 3.13 | 2.96 | **2.81** |
| **Ave Execution Distribution** | | | | | | |
| HGSO | 16.66% | 26.45% | 18.97% | 19.26% | 20.06% | 32.86% |
| JA | 36.67% | 20.57% | 22.64% | 22.84% | 19.47% | 16.90% |
| STOA | 16.67% | 4.94% | 18.28% | 19.75% | 20.57% | 16.90% |
| BOA | 10.00% | 21.49% | 21.26% | 16.30% | 20.30% | 16.44% |
| OSA | 20.00% | 26.55% | 18.85% | 21.85% | 19.60% | 16.90% |

Table 5. Effect of Cluster Size based on DBLP Data Set for 30, 60, and 90 required skills

| | HHGSO | | | | | |
|---|---|---|---|---|---|---|
| | Cluster Size=1 | Cluster Size=2 | Cluster Size=3 | Cluster Size=4 | Cluster Size=5 | Cluster Size=6 |
| **30 required skills** | | | | | | |
| **Metrics Measurement** | | | | | | |
| Best Cost | 342.63 | 342.50 | 318.49 | 342.13 | **317.16** | 342.65 |
| No of Team Members | 27 | 27 | **26** | 27 | **26** | 27 |
| Ave Cost | 353.86 | 343.02 | 340.15 | 352.44 | **340.10** | 351.50 |
| Ave Time (sec) | 468.74 | 443.50 | **435.51** | 508.20 | 439.87 | 437.73 |
| **Ave Execution Distribution** | | | | | | |
| HGSO | 20.00% | 25.27% | 18.26% | 24.26% | 19.23% | 32.59% |
| JA | 1.71% | 25.40% | 21.99% | 23.55% | 20.27% | 16.87% |
| STOA | 1.71% | 7.18% | 17.34% | 14.38% | 19.94% | 16.85% |
| BOA | 10.29% | 4.78% | 23.63% | 22.33% | 20.23% | 16.77% |
| OSA | 66.29% | 37.37% | 18.78% | 15.48% | 20.33% | 16.92% |
| **60 required skills** | | | | | | |
| **Metrics Measurement** | | | | | | |
| Best Cost | 1464.35 | 1518.86 | **1463.43** | 1467.23 | 1469.29 | 1466.98 |
| No of Team Members | **55** | 56 | **55** | **55** | **55** | **55** |
| Ave Cost | 1518.43 | 1538.00 | 1520.11 | 1521.91 | **1502.70** | 1538.10 |
| Ave Time (sec) | **418.53** | 495.69 | 444.20 | 426.68 | 472.26 | 485.51 |
| **Ave Execution Distribution** | | | | | | |
| HGSO | 42.06% | 34.92% | 15.80% | 16.13% | 19.64% | 32.45% |
| JA | 6.35% | 13.61% | 19.82% | 21.63% | 20.22% | 16.84% |
| STOA | 38.89% | 25.44% | 30.04% | 23.99% | 20.31% | 16.91% |
| BOA | 0.80% | 13.62% | 19.42% | 23.74% | 20.25% | 16.95% |
| OSA | 11.90% | 12.41% | 14.92% | 14.51% | 19.58% | 16.85% |
| **90 required skills** | | | | | | |
| **Metrics Measurement** | | | | | | |
| Best Cost | 3187.90 | 3264.27 | 3194.46 | 3199.22 | **3035.01** | 3182.78 |
| No of Team Members | 81 | 82 | 81 | 81 | **79** | 81 |
| Ave Cost | 3243.30 | 3383.86 | 3352.99 | 3380.93 | **3327.03** | 3379.78 |
| Ave Time (sec) | 406.12 | 460.71 | **397.94** | 443.88 | 421.22 | 457.33 |
| **Ave Execution Distribution** | | | | | | |
| HGSO | 7.94% | 24.68% | 25.57% | 14.98% | 20.31% | 33.54% |
| JA | 10.32% | 40.12% | 19.53% | 23.32% | 20.23% | 16.22% |
| STOA | 66.67% | 5.13% | 19.90% | 23.39% | 19.44% | 16.32% |
| BOA | 5.56% | 8.33% | 31.77% | 22.57% | 20.31% | 16.95% |
| OSA | 9.51% | 21.74% | 3.23% | 15.74% | 19.71% | 16.97% |

Concerning the second issue, the average execution distribution for each participating constituent algorithms for cluster size=1 until 6 are referred to in Table 4 for IMDB and in Table 5 for DBLP. Conveniently, the average execution distributions are plotted as cascaded bar charts in Figure 6(a) till 6(c) and 7(a) till 7(c), respectively. Firstly, from Figure 6 and 7, all algorithms do have a chance to participate in the search process.

Considering HHGSO with cluster size =1 till 3, there is no clear pattern of preferences to any particular meta-heuristic algorithm. This is expected as there are randomized many-to-one-assignment of meta-heuristic algorithm to cluster(s). The pattern seems to change in the case of HHGSO with cluster size=4, the distribution of each algorithm is nearly even owing to the fact that the mapping is now many-to-many (i.e., algorithms to clusters).

It is obvious that HHGSO with cluster size=5 is expected to have 20% execution distribution each. This observation does not exactly materialize in the average distribution results (see Table 4 and 5). The main reason is that search execution may stop much earlier than maximum iteration, that is, when maximum fitness evaluation is reached. In fact, execution may stop in between any cluster execution (i.e., based on the maximum fitness evaluation). Therefore, some algorithm may have less execution than others (resulting into slightly less average execution). Another obvious observation relates to the preference to HHGSO when the cluster size=6. Here, there is one extra cluster more than the defined 5 participating constituent algorithms. This extra cluster by default is assigned to HGSO guaranteeing 2 cluster mapping out of a total 6 clusters. Whereas, other algorithm share only 1 mapping out of 6 clusters. As more and more clusters are defined and with fixed numbers of participating constituent algorithms, it is expected that the cluster mapping will be more and more biased toward HGSO.

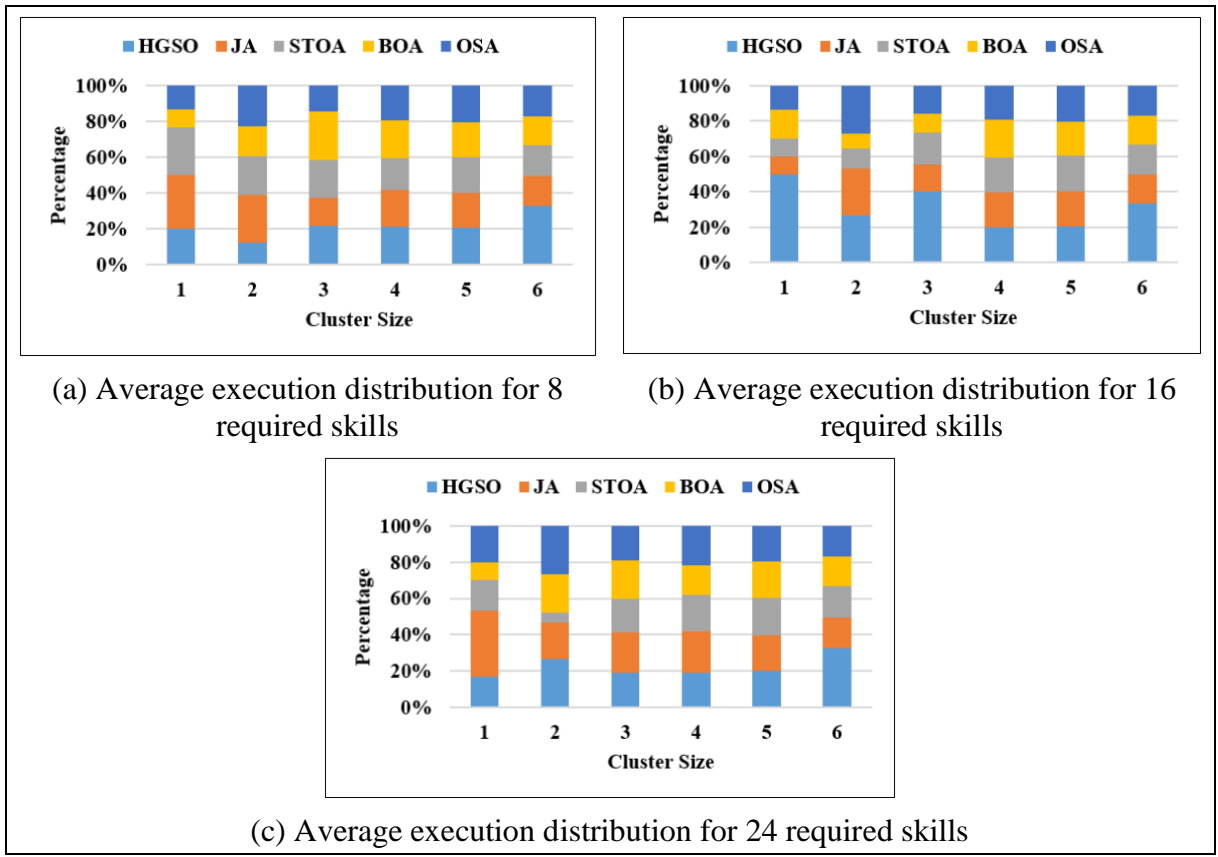

Figure 6. Average Execution Distribution based on IMDB Data Set for 8, 16, and 24 required skills

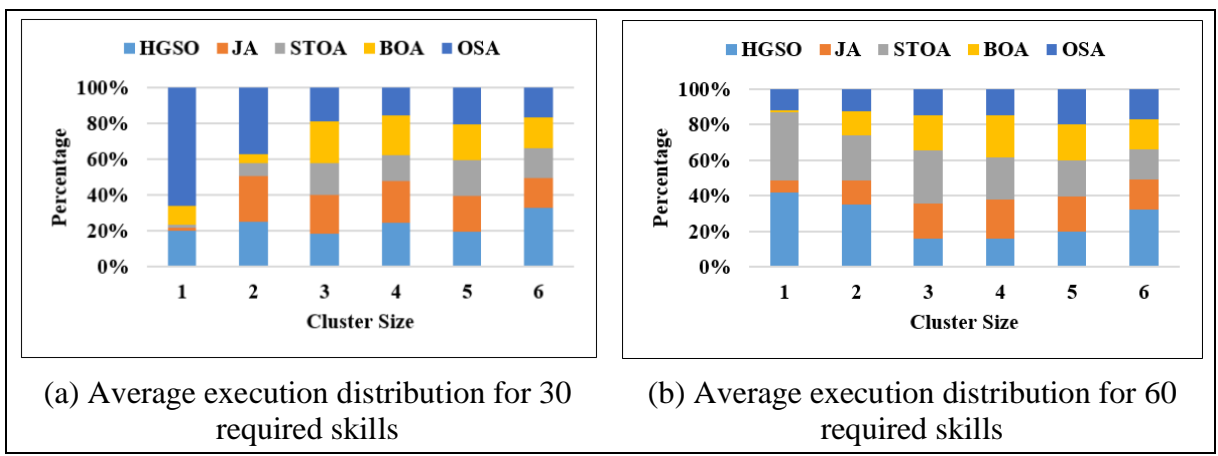

(a) Average execution distribution for 30 required skills

(b) Average execution distribution for 60 required skills

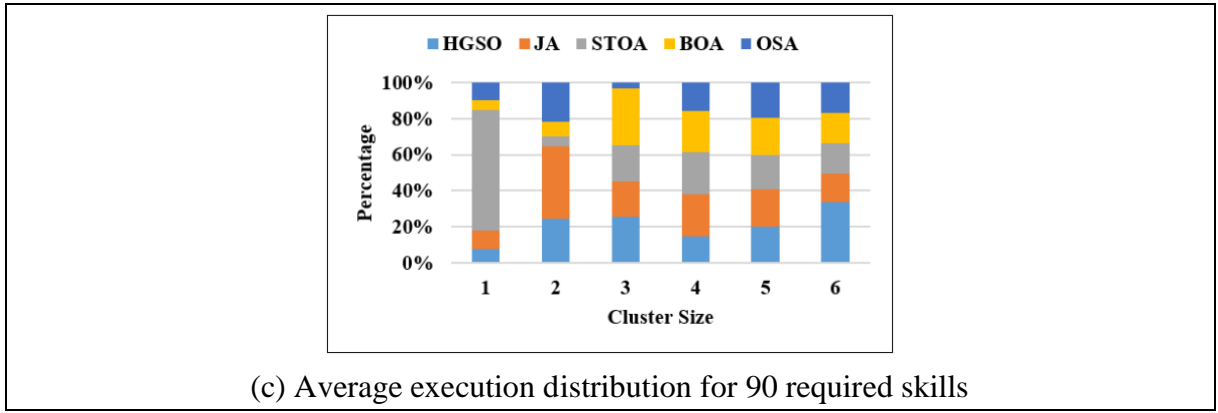

(c) Average execution distribution for 90 required skills

Figure 7. Average Execution Distribution based on DBLP Data Set for 30, 60, and 90 required skills

- **RQ3: How good is the performance of HHGSO against its cooperative MASP-HLH counter parts (i.e., hyper-heuristic algorithms)?**

Results from Table 6 till 12 indicate that the performance of HHGSO is at par with Fuzzy inference Selection hyper-heuristic algorithm (as another form of MASP-HLH derived from hyper-heuristic family of algorithms). Reporting of many of new best in the literature, outperforming Fuzzy Inference Selection hyper-heuristic algorithm is a challenge. Despite such challenge, HHGSO manage to get new overall best for *CA$_2$ (N; 2, 10$^{10}$)* and *CA$_5$ (N; 3, 10$^6$)* (in Table 6) as well as C*A (N; 4, v$^6$)* (in Table 12). The rest of the cells, more often than not, HHGSO only manage to equal the best results from Fuzzy Inference Selection hyper-heuristic algorithm. On the positive note, HHGSO gives better average test size than FIS in cells throughout Table 6 till Table 12 indicating its consistent performance. Putting Fuzzy Inference Selection hyper-heuristic aside, HHGSO outperforms all other compared hyper-heuristic algorithms. Improvement Selection Rules and Modified Choice function comes in joint third and closely followed by Exponential Monte Carlo with Counter.

Table 6. Size Performance for Selected *CAs*

| CA | Cooperative MASP-HLH | | | | | | | | | |
|---|---|---|---|---|---|---|---|---|---|---|
| | Exponential Monte Carlo with Counter | | Modified Choice Function | | Improvement Selection Rules | | Fuzzy Inference Selection | | Hybrid HGSO | |
| | Best | Ave | Best | Ave | Best | Ave | Best | Ave | Best | Ave |
| $CA_1 (N; 2, 3^{13})$ | 18 | 19.05 | 18 | 19.45 | 18 | 18.90 | **17** | **18.65** | **17** | 18.80 |
| $CA_2 (N; 2, 10^{10})$ | 155 | 157.20 | 157 | 172.05 | 156 | 157.35 | 153 | 157.10 | **150** | **154.60** |
| $CA_3 (N; 3, 3^6)$ | **33** | 38.85 | **33** | 38.90 | **33** | 37.75 | **33** | 38.20 | **33** | **38.00** |
| $CA_4 (N; 3, 6^6)$ | 323 | 326.70 | 323 | 327.40 | **322** | 326.20 | 323 | **326.15** | 326 | 328.90 |
| $CA_5 (N; 3, 10^6)$ | 1485 | 1496.50 | 1483 | 1499.25 | 1482 | 1486.80 | 1481 | 1486.20 | **1473** | **1478.90** |
| $CA_6 (N; 3, 5^2 4^2 3^2)$ | **100** | 107.35 | **100** | 113.20 | **100** | 105.55 | **100** | 105.95 | **100** | **105.30** |

Table 7. Size Performance for *CA (N; 2, $3^k$)*

| K | Cooperative MASP-HLH | | | | | | | | | |
|---|---|---|---|---|---|---|---|---|---|---|
| | Exponential Monte Carlo with Counter | | Modified Choice Function | | Improvement Selection Rules | | Fuzzy Inference Selection | | Hybrid HGSO | |
| | Best | Ave | Best | Ave | Best | Ave | Best | Ave | Best | Ave |
| 3 | **9** | 9.83 | **9** | 9.70 | **9** | 9.90 | **9** | **9.67** | **9** | 9.81 |
| 4 | **9** | **9.00** | **9** | **9.00** | **9** | **9.00** | **9** | **9.00** | **9** | **9.00** |
| 5 | **11** | 11.24 | **11** | 11.30 | **11** | 11.30 | **11** | **11.23** | **11** | **11.23** |
| 6 | 14 | 14.27 | **13** | 14.36 | **13** | 14.46 | **13** | 14.03 | **13** | **14.00** |
| 7 | 15 | **15.07** | 15 | 15.23 | 15 | 15.10 | **14** | **15.07** | 15 | 15.10 |
| 8 | **15** | 15.77 | **15** | 16.16 | **15** | 15.90 | **15** | 15.79 | **15** | **15.75** |
| 9 | **15** | 16.23 | **15** | 16.43 | **15** | 16.10 | **15** | **15.97** | **15** | 16.05 |
| 10 | **16** | 17.10 | **16** | 17.20 | **16** | 17.50 | **16** | 17.03 | **16** | **17.01** |
| 11 | 17 | 18.90 | 18 | 18.50 | 17 | 18.30 | **16** | **17.45** | **16** | 17.60 |
| 12 | **16** | 17.96 | 17 | 18.29 | 17 | 18.40 | **16** | 17.80 | **16** | **17.79** |

Table 8. Size Performance for *CA (N; 3, $3^k$)*

| K | Cooperative MASP-HLH | | | | | | | | | |
|---|---|---|---|---|---|---|---|---|---|---|
| | Exponential Monte Carlo with Counter | | Modified Choice Function | | Improvement Selection Rules | | Fuzzy Inference Selection | | Hybrid HGSO | |
| | Best | Ave | Best | Ave | Best | Ave | Best | Ave | Best | Ave |
| 4 | **27** | 28.83 | **27** | 29.20 | **27** | 30.06 | **27** | 27.23 | **27** | **27.10** |
| 5 | 39 | 41.47 | 38 | 41.40 | 39 | 41.60 | **37** | **41.30** | 39 | 41.55 |
| 6 | **33** | 38.63 | **33** | 38.37 | **33** | 38.47 | **33** | 36.77 | **33** | **36.70** |
| 7 | 49 | 50.46 | 49 | 50.50 | 49 | 50.47 | **48** | **50.40** | **48** | **50.40** |
| 8 | **52** | 53.27 | **52** | 53.93 | **52** | 53.27 | 53 | 53.40 | **52** | **53.10** |
| 9 | **56** | 57.79 | 57 | 58.07 | **56** | 57.87 | **56** | 57.77 | **56** | **57.60** |
| 10 | **59** | 61.17 | 60 | 60.77 | 60 | 60.10 | **59** | 61.03 | **59** | **61.00** |
| 11 | **63** | 63.87 | 64 | 65.27 | **63** | 63.67 | **63** | **63.53** | **63** | **63.53** |
| 12 | **65** | 67.61 | 66 | 68.13 | **65** | 66.93 | **65** | **66.13** | **65** | 66.20 |

Table 9. Size Performance for *CA (N; 4, 3$^k$)*

| k | Cooperative MASP-HLH | | | | | | | | | |
|---|---|---|---|---|---|---|---|---|---|---|
| | Exponential Monte Carlo with Counter | | Modified Choice Function | | Improvement Selection Rules | | Fuzzy Inference Selection | | Hybrid HGSO | |
| | Best | Ave | Best | Ave | Best | Ave | Best | Ave | Best | Ave |
| 5 | **81** | **84.23** | **81** | 89.07 | **81** | 88.27 | **81** | 87.27 | **81** | 87.11 |
| 6 | 130 | 133.33 | **129** | 133.83 | **129** | 134.17 | **129** | 134.10 | 130 | **133.10** |
| 7 | 149 | 154.27 | 151 | 155.17 | **147** | 153.53 | **147** | 153.90 | **147** | 153.67 |
| 8 | 172 | 174.96 | 173 | 175.47 | **171** | 174.83 | **171** | 174.47 | **171** | 174.41 |
| 9 | 160 | **187.87** | 142 | 190.53 | 171 | 190.33 | 159 | 189.47 | 178 | 189.05 |
| 10 | 206 | 209.00 | **205** | 208.83 | 206 | 208.77 | 206 | 208.67 | **205** | 208.22 |
| 11 | **221** | 224.67 | 222 | 226.13 | **221** | 224.33 | **221** | 223.13 | **221** | 223.13 |
| 12 | 237 | 238.51 | 237 | 239.21 | 236 | 238.11 | **235** | **237.43** | **235** | 237.60 |

Table 10. Size Performance for *CA (N; 2, v$^7$)*

| v | Cooperative MASP-HLH | | | | | | | | | |
|---|---|---|---|---|---|---|---|---|---|---|
| | Exponential Monte Carlo with Counter | | Modified Choice Function | | Improvement Selection Rules | | Fuzzy Inference Selection | | Hybrid HGSO | |
| | Best | Ave | Best | Ave | Best | Ave | Best | Ave | Best | Ave |
| 2 | **7** | **7.00** | 7 | **7.00** | 7 | **7.00** | 7 | **7.00** | 7 | **7.00** |
| 3 | 15 | 15.13 | 15 | 15.13 | 15 | 15.17 | **14** | **15.00** | **14** | **15.00** |
| 4 | 24 | 25.07 | 24 | 25.47 | **23** | 25.00 | 24 | 24.87 | 24 | **24.40** |
| 5 | **34** | 35.83 | **34** | 36.63 | **34** | 35.90 | **34** | 35.70 | **34** | **35.40** |
| 6 | 48 | 49.00 | 48 | 49.67 | **47** | 49.51 | **47** | 48.75 | **47** | **48.40** |
| 7 | **64** | 65.93 | **64** | 66.85 | **64** | 66.25 | **64** | 65.65 | **64** | **65.45** |

Table 11. Size Performance for *CA (N; 3, v$^7$)*

| v | Cooperative MASP-HLH | | | | | | | | | |
|---|---|---|---|---|---|---|---|---|---|---|
| | Exponential Monte Carlo with Counter | | Modified Choice Function | | Improvement Selection Rules | | Fuzzy Inference Selection | | Hybrid HGSO | |
| | Best | Ave | Best | Ave | Best | Ave | Best | Ave | Best | Ave |
| 2 | 14 | 15.17 | 15 | 15.17 | 15 | 15.20 | **12** | **15.00** | 15 | 15.20 |
| 3 | 49 | 50.60 | **48** | 50.53 | **48** | 50.57 | **48** | 50.47 | **48** | **50.34** |
| 4 | 113 | 115.70 | 114 | 115.07 | 113 | 115.37 | **112** | 114.90 | **112** | **114.70** |
| 5 | 217 | 220.37 | **215** | 219.00 | 216 | 218.65 | 216 | **218.60** | 217 | 218.85 |
| 6 | **365** | 373.91 | 369 | 374.43 | 365 | 373.51 | 366 | 370.20 | **365** | **370.01** |
| 7 | **575** | 579.00 | **575** | 580.91 | **575** | 579.75 | **575** | 577.80 | **575** | 578.20 |

Table 12. Size Performance for *CA (N; 4, v$^7$)*

| v | Cooperative MASP-HLH | | | | | | | | | |
|---|---|---|---|---|---|---|---|---|---|---|
| | Exponential Monte Carlo with Counter | | Modified Choice Function | | Improvement Selection Rules | | Fuzzy Inference Selection | | Hybrid HGSO | |
| | Best | Ave | Best | Ave | Best | Ave | Best | Ave | Best | Ave |
| 2 | 31 | 32.23 | 31 | 31.80 | 31 | 32.27 | **26** | **31.67** | 31 | 31.86 |
| 3 | 151 | 155.30 | 151 | 155.27 | 151 | 154.53 | **150** | 154.40 | **150** | **154.29** |
| 4 | **479** | 484.83 | **479** | 485.17 | **479** | 484.00 | 480 | 484.00 | **479** | **483.88** |
| 5 | 1156 | 1161.47 | **1151** | **1160.03** | 1154 | 1162.43 | 1154 | 1161.03 | 1156 | 1162.30 |
| 6 | 2348 | 2364.23 | 2353 | 2369.91 | 2352 | 2367.11 | 2349 | 2363.11 | **2347** | **2362.99** |
| 7 | 4294 | 4311.10 | 4295 | 4312.70 | 4295 | 4311.90 | **4293** | **4310.54** | 4294 | 4311.30 |

- **RQ4: Is there any overhead in terms time performance penalty of HHGSO implementation as compared to its constituent algorithms?**

Time performance of HHGSO relates to its computational complexity metric. For this purpose, the Big O notation is used. Complexity is dependent on the number of search agents (*n*), the number of dimensions (*d*), the number of maximum iteration (*Max$_{iter}$*) and the fitness function evaluation (*c*).

As HHGSO consists of five meta-heuristic algorithms (i.e., HGSO, JA, STOA, BOA, OSA), its complexity *O(HHGSO)* is a combination of *O(HGSO)+O(JA)+O(STOA)+O(BOA)+O(OSA)*. To analyze the overall time complexity, there is a need to analyze the contribution of each algorithm. With the cluster size = *N_size*, the time complexity of each algorithm is:

$$O(HGSO) = O(fitness\ function\ evaluation)+O(agent\ update\ in\ memory)+O(dimension\ update)$$
$$= O(Max_{iter}\times c\times n/N\_size + Max_{iter}\times n/N\_size + Max_{iter}\times d\times n/N\_size)$$
$$\cong O(Max_{iter}\times c\times n/N\_size + Max_{iter}\times d\times n/N\_size) \quad (37)$$

$$O(JA) = O(fitness\ function\ evaluation)+O(agent\ update\ in\ memory)+O(dimension\ update)$$
$$= O(Max_{iter}\times c\times n/N\_size + Max_{iter}\times n/N\_size + Max_{iter}\times d\times n/N\_size)$$
$$\cong O(Max_{iter}\times c\times n/N\_size + Max_{iter}\times d\times n/N\_size) \quad (38)$$

$$O(STOA) = O(fitness\ function\ evaluation)+O(agent\ update\ in\ memory)+O(dimension\ update)$$
$$= O(Max_{iter}\times c\times n/N\_size + Max_{iter}\times n/N\_size + Max_{iter}\times d\times n/N\_size)$$
$$\cong O(Max_{iter}\times c\times n/N\_size + Max_{iter}\times d\times n/N\_size) \quad (39)$$

$$O(BOA) = O(fitness\ function\ evaluation)+O(agent\ update\ in\ memory)$$
$$+O(new\ fragrance\ generation\ loop)+O(dimension\ update)$$
$$= O(Max_{iter}\times c\times n/N\_size + Max_{iter}\times n/N\_size + Max_{iter}\times n/N\_size + Max_{iter}\times d\times n/N\_size)$$
$$\cong O(Max_{iter}\times c\times n/N\_size + Max_{iter}\times d\times n/N\_size) \quad (40)$$

$$O(OSA) = O(fitness\ function\ evaluation)+O(agent\ update\ in\ memory)+O(dimension\ update)$$
$$= O(Max_{iter}\times c\times n/N\_size + Max_{iter}\times n/N\_size + Max_{iter}\times d\times n/N\_size)$$
$$\cong O(Max_{iter}\times c\times n/N\_size + Max_{iter}\times d\times n/N\_size) \quad (41)$$

Combining each algorithm's contribution, and generalizing with the number of participating meta-heuristic algorithms (*S*):

$$O(HHGSO) \cong S\times O(Max_{iter}\times c\times n/N\_size + Max_{iter}\times d\times n/N\_size) \quad (42)$$

Consider the limit when *N_size =1* and *S=1*, then:

$$O(HHGSO) \cong O(Max_{iter}\times c\times n + Max_{iter}\times d\times n) \quad (43)$$

As can be seen from the aforementioned derivations (Eq. (37) till Eq. (43)), the general time complexity of HHGSO is similar to each of its constituent. The time complexity multiplier depends on the scaling factor = the number of participating algorithms.

- **RQ5: How generalized can HHGSO implementation be for solving general optimization problems?**

We have implemented and subjected HHGSO to two Search based Software Engineering problems involving the team formation and the combinatorial t-way test suite generation. Our experimental results give clear indication that HHGSO approach can produce competitive results. Given that the two problems are domain specific in nature, the fact that HHGSO is able to optimize their problem specific objective functions speaks volume of its applicability to other optimization problems as well.

Generally, HHGSO tends to outperform general meta-heuristic algorithms. This could be due to the fact that HHGSO has the benefit of being able to utilize more than one meta-heuristic algorithm to perform the search process. Furthermore, the dynamic meta-heuristic-algorithm-to-cluster-mapping usefully promotes exploration diversity whereby the displacement of any search agent in a cluster is not necessarily fixed to a particular type of update operator from one meta-heuristic algorithm only. In fact, such feature could also be useful as a way to avoid entrapment in local optima.

Finally, while our hybridization utilizes a combination of four specific meta-heuristic algorithms (i.e., Jaya, Sooty Tern Optimization Algorithm, Butterfly Optimization Algorithm and Owl Search Algorithm), our work is still sufficiently general given its adaptability and ease of use. In fact, any known meta-heuristic algorithms can be adopted as part of our HHGSO implementation.

## 7. Discussion

Reflecting on research questions given earlier, the usefulness of our approach can be debated further.

Arguably, our implementation of HHGSO reveals two subtle properties. The first property relates to the meta-heuristic-algorithm-to-cluster mapping. To be specific, meta-heuristic-algorithm-to-cluster-mapping is dynamically mapped based a single population (i.e., with more than search agents). At any instance of iteration, the cluster keeps its local best independently of the assigned meta-heuristic algorithm similar to the original HGSO implementation. Figure 6 depicts visual execution of HHGSO with 12 populations at any iteration = J and K.

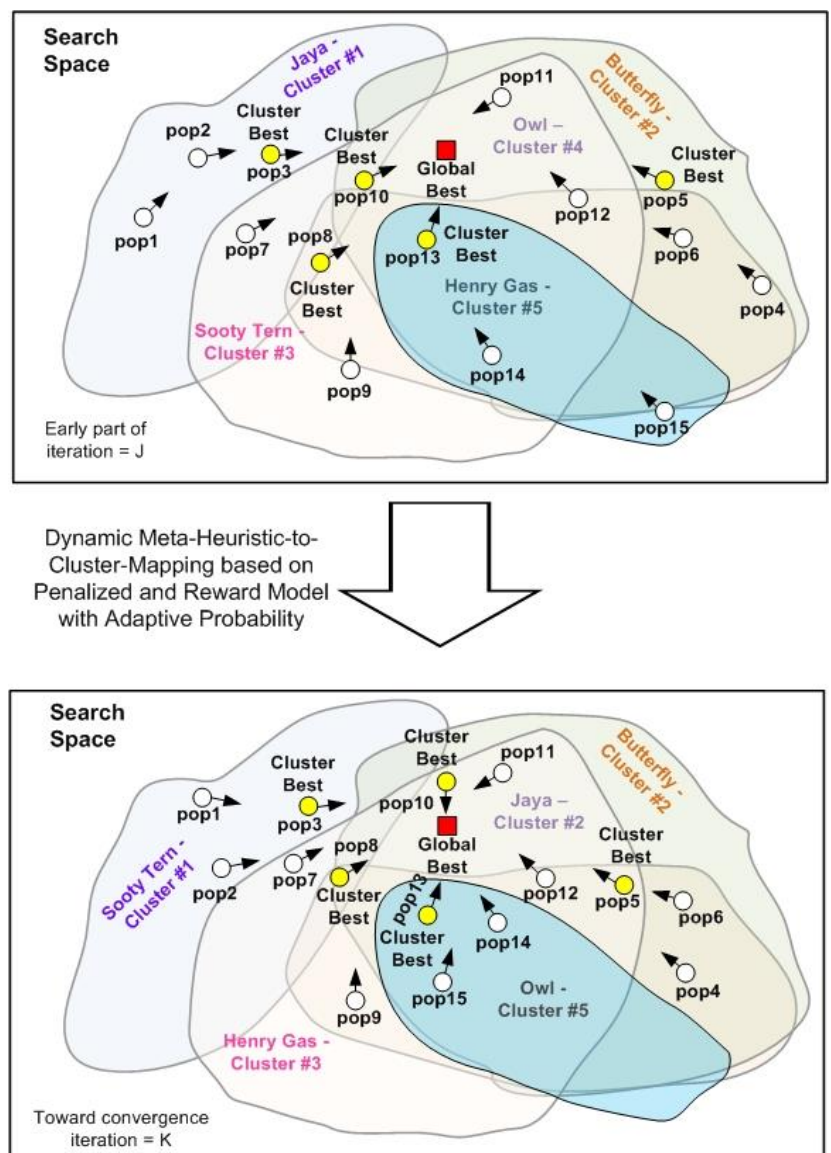

Figure 6. Visual Representation of HHGSO Execution

The second property relates to the flexibility of HHGSO implementation as compared to its predecessor. Referring to Figure 2 given earlier, HHGSO in the current form can be categorized as cooperative MASP-HLH (as all the algorithms works together as a unit in unpredictable sequence and is non-intersecting to the host algorithm features). If the adaptive probability associated with the proposed co-operative penalized and reward model is removed from the implementation, the resulting HHGSO can be categorized as relay based MASP-HLH. Taking the discussion further, consider the normal scenario when the cluster size is the same as the total list of participating meta-heuristic algorithms. Here, there will always be a unique one-to-one dynamic assignment of meta-heuristic-algorithm-to-cluster-mapping at any iteration. In the case where the cluster size is less than the list of participating meta-heuristic algorithms, the dynamic assignment of meta-heuristic-algorithm-to-cluster-mapping will be randomly decided among those algorithms. As such, at any particular iteration, there are algorithms that will not be participating in the meta-heuristic-algorithm-to-cluster-mapping (i.e., as cluster size < list of participating algorithms). Nonetheless, in the end, all the participating meta-heuristic algorithms do have a chance to run at least once during any of the iteration. This flexibility allows HHGSO to conveniently ride on many participating meta-heuristic algorithms without rigidly tied to specific algorithm. For example, one can define 30 participating meta-heuristic algorithms with just 3 defined clusters. With the 3 defined clusters, there will be 3 defined meta-heuristic-algorithm-to-cluster-mappings in each iteration. Here, the 3 defined meta-heuristic-algorithm-to-cluster-mappings have the luxury to randomly adopt any 3 participating meta-heuristic algorithms based on penalized and reward probability model.

In one extreme case, even with only 1 defined cluster, we can still have cooperative MASP-HLH (i.e., considering many single algorithms in use although with just 1 cluster of population). Similarly, at the other extreme, we can also have more clusters than the number of participating algorithms. In such a case, HHGSO will have more HGSO-to-cluster-mappings than other meta-heuristic-algorithm-to-cluster mappings on the virtue of being the host algorithm. This aforementioned flexibility is unique to HHGSO implementation and is not found in any current hybridization scheme in the literature.

## 8. Concluding Remarks

In this paper, we have presented a new form of hybridization based on HGSO, termed HHGSO. Taking HGSO as the host algorithm, HHGSO rides on four recently developed meta-heuristic algorithms including Jaya Algorithm (JA), Sooty Tern Optimization Algorithm (STOA), Butterfly Optimization Algorithm (BOA) and Owl Search Algorithm (OSA). As part of our analysis on the related work, we have categorized existing hybridization based on their population implementations considering the level of participating algorithms' integration (i.e. low level versus high level) as well as considering how their cooperation take place (i.e. relay versus cooperative).

The main feature of HHGSO is twofold. Firstly, HHGSO divides the population into clusters with dynamic meta-heuristic-algorithms-to-cluster mapping. Secondly, HHGSO uses an adaptive probability based on penalized and reward model to switch between cluster mappings. The two features working together gives HHGSO an edge over other approaches.

In order to conduct extensive evaluation, we have subjected HHGSO to two Search based Software Engineering problems. We have concluded that HHGSO is sufficiently general and can be applicable to other optimization problems.

As the scope for future work, we are looking to integrate HHGSO with other combinations of meta-heuristic algorithms. In doing so, we are also interested to apply HHGSO to other domain specific optimization problems (e.g. travelling salesman, bin-packing, and vehicle routing problems) owing to its promising performance.

## Acknowledgement

The work reported in this paper is funded by Fundamental Research Grant from the Ministry of Higher Education Malaysia titled: A Reinforcement Learning Sine Cosine based Strategy for Combinatorial Test Suite Generation (grant no: RDU170103).